\documentclass[pdflatex,sn-mathphys-ay,iicol]{sn-jnl}

\usepackage{graphicx}%
\usepackage{multirow}%
\usepackage{amsmath}%
\usepackage{amssymb}%
\usepackage{amsfonts}%
\usepackage{amsthm}%
\usepackage{mathrsfs}%
\usepackage[title]{appendix}%
\usepackage{xcolor}%
\usepackage{textcomp}%
\usepackage{manyfoot}%
\usepackage{booktabs}%
\usepackage[ruled,vlined,linesnumbered]{algorithm2e}
\SetKwInput{KwReq}{Require}
\SetKw{KwRet}{return}

\usepackage{listings}%
\usepackage{bm}%
\usepackage{array}%
\usepackage{color}
\usepackage[accsupp]{axessibility}
\usepackage{diagbox}
\usepackage{adjustbox}
\usepackage{float}

\usepackage{cuted}   

\usepackage{makecell}
\usepackage{caption}  

\usepackage[T1]{fontenc}
\usepackage[utf8]{inputenc}

\theoremstyle{thmstyleone}%
\theoremstyle{thmstyletwo}%

\theoremstyle{thmstylethree}%

\begin{document}
\title[Article Title]{Privacy-Preserving Semantic Segmentation from Ultra-Low-Resolution RGB Inputs}


\author*[1,2,3]{\fnm{Xuying} \sur{Huang}}\email{huang@cs.uni-bonn.de}

\author[1,2,3]{\fnm{Sicong} \sur{Pan}}\email{pan@cs.uni-bonn.de}

\author[1]{\fnm{Olga} \sur{Zatsarynna}}\email{zatsarynna@iai.uni-bonn.de}

\author[1,2,3]{\fnm{Juergen} \sur{Gall}}\email{gall@iai.uni-bonn.de}

\author[1,2,3]{\fnm{Maren} \sur{Bennewitz}}\email{maren@cs.uni-bonn.de}

\affil[1]{\orgdiv{University of Bonn}, \country{Germany}}
\affil[2]{\orgdiv{Lamarr Institute for Machine Learning and Artificial Intelligence}, \country{Germany}}
\affil[3]{\orgdiv{Center for Robotics}, \country{Germany}}

\abstract{
RGB-based semantic segmentation has become a mainstream approach for visual perception and is widely applied in a variety of downstream tasks. 
However, existing methods typically rely on high-resolution RGB inputs, which may expose sensitive visual content in privacy-critical environments. 
Ultra-low-resolution RGB sensing suppresses sensitive information directly during image acquisition, making it an attractive privacy-preserving alternative.
Nevertheless, recovering semantic segmentation from ultra-low-resolution RGB inputs remains highly challenging due to severe visual degradation.
In this work, we introduce a novel fully joint-learning framework to mitigate the optimization conflicts exacerbated by visual degradation for ultra-low-resolution semantic segmentation.
Experiments demonstrate that our method outperforms representative baselines in semantic segmentation performance and our ultra-low-resolution RGB input achieves a favorable trade-off between privacy preservation and semantic segmentation performance.
We deploy our privacy-preserving semantic segmentation method in a real-world robotic object-goal navigation task, demonstrating successful downstream task execution even under severe visual degradation.
}

\keywords{Privacy preservation, Semantic segmentation, Ultra-low-resolution RGB, Object-goal navigation}

\maketitle

\begin{figure*}[t] 
    \centering
    \includegraphics[width=1.0\textwidth]{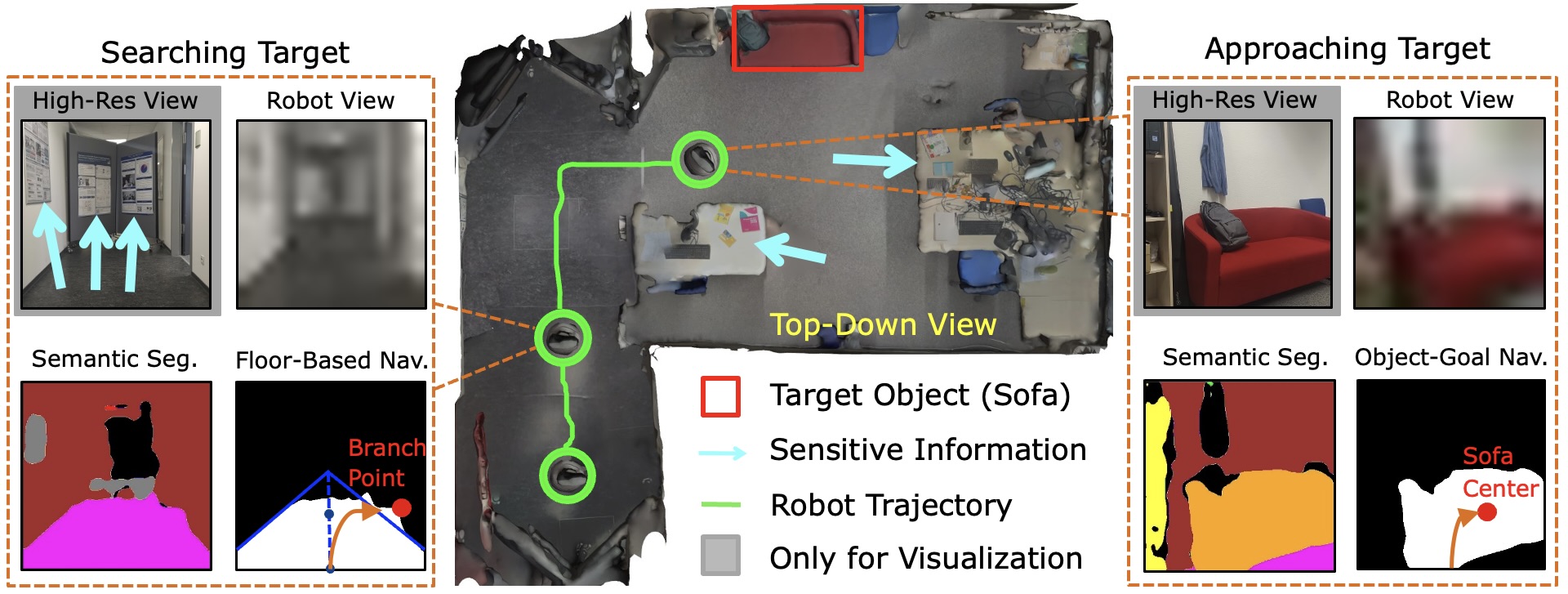} 
    
    \caption{
    Illustration of a real-world application scenario of our privacy-preserving ultra-low-resolution semantic segmentation framework deployed for object-goal navigation. 
    The center panel shows a top-down view of the robot's trajectory (green line) toward the target (red box) in an environment containing privacy-sensitive content (cyan arrows).
    The High-Res Views show full-resolution RGB images for visualization only, while the Robot Views display the actual robot input: ultra-low-resolution ($16 \times 16$) monocular RGB images.
    On the left, when the target has not yet been found, the robot performs floor-based navigation by following waypoints (blue dots) along the floor centerline (blue dashed line), turning at branch points that may lead to new rooms.
    On the right, once the target has been found, the robot switches to goal navigation, gradually approaching the target.
    As can be seen, our method achieves plausible semantic segmentation from ultra-low-resolution RGBs while benefiting privacy-preserving robotic task execution. 
    }
    \label{fig:teaser}
    \vspace{-0.4cm}
\end{figure*}

\section{Introduction}\label{sec1}
Semantic segmentation from RGB images has emerged as a fundamental technique in computer vision, enabling dense pixel-level understanding of visual scenes~\citep{Shin2024neurips}.
It has been widely deployed across various domains, such as autonomous driving~\citep{Zhao2024cvpr}, medical imaging~\citep{Rahman2024cvpr}, and in particular mobile robotics~\citep{Menon2025iros}.
Semantic segmentation provides a critical environmental context for mobile robots to facilitate downstream tasks. 
A representative application is object-goal navigation, where a robot is asked to find and approach a target object, such as a chair, desk, monitor, or sofa~\citep{yokoyama2024icra}.

However, existing RGB-based semantic segmentation approaches typically assume the availability of high-resolution (HR) RGB inputs to extract fine-grained visual details~\citep{Wang2020tpami}.
While beneficial for perception, relying on HR imagery in real-world deployments introduces potential privacy risks. 
For example, mobile robots move into sensitive everyday environments like homes, offices, and healthcare facilities, their onboard cameras may inadvertently capture identifiable individuals, personal belongings, computer screens, and confidential materials\footnote{\url{https://www.theguardian.com/uk/2012/nov/20/prince-william-photos-mod-passwords}}.

To mitigate such risks, existing works have extensively explored post-capture privacy protection like sanitization~\citep{Dufaux2008tcsvt}, de-identification~\citep{Wen2023iccv}, or substituting RGB images with privacy-friendly representations~\citep{Moon2023aaai}.
However, these methods either depend on HR RGB inputs or require the initial capture of HR frames with temporary buffering of the original images, thereby still introducing privacy risks in network-connected deployments.
In particular, this buffering introduces an attack surface: if the device is compromised during this short window, sensitive visual or textual information could be exfiltrated~\citep{dritsas202fi, sasi2024jii}.
Compared to post-processing, one promising direction is to enforce privacy at capture time by minimizing data acquisition through ultra-low-resolution~(ULR) imaging. 
Prior studies have shown that ULR RGB, such as $15 \times 15$~\citep{kim2019iros} and $16 \times 12$~\citep{Ryoo2018aaai}, can effectively suppress identifiable visual details.
A recent user study~\citep{huang2025arxiv} also suggests that resolution around $16 \times 16$ is privacy-preferable. 
Building on these observations, we therefore prefer ULR~($16 \times 16$) paradigm for privacy protection. 

However, semantic segmentation from ULR RGB inputs is highly challenging. 
Despite its privacy benefits, the severe degradation of visual information hinders the performance of semantic segmentation methods.
Existing approaches mainly fall into three categories: (1)~pretrained pipelines~\citep{caputa2024life}, which use pretrained segmentation alone or cascaded pretrained SR and segmentation networks; (2)~mono-module training~\citep{frizza2022cviu}, which updates only the SR or segmentation module while freezing the other; and (3)~naive fully joint learning of both modules~\citep{pereira2020lagirs}. 
Nevertheless, none of these designs yields satisfactory semantic segmentation performance.

Therefore, to overcome these limitations, we introduce a novel joint-learning framework that incorporates a segmentation-aware discriminator and an agglomerative feature extractor to mitigate the optimization conflicts exacerbated by visual degradation. This architecture enables effective recovery of semantic segmentation from ULR RGB images. 
For simplicity, we refer to our pipeline that recovers plausible semantic segmentation from ULR RGB images as ULR semantic segmentation.
To the best of our knowledge, this work is the first to enable ULR semantic segmentation for privacy-preserving downstream tasks.
Fig.~\ref{fig:teaser} illustrates a real-world object-goal navigation application scenario of our method.
Our contributions are the following:
\begin{itemize}
    \item \textbf{Joint-learning framework:} We introduce a novel joint-learning architecture, outperforming representative baselines in terms of ULR semantic segmentation performance under extreme visual degradation.
    \item \textbf{Joint-learning conflict analysis:} We demonstrate that explicitly mitigating the conflict between super-resolution and segmentation objectives is critical for effective ULR semantic segmentation.
    \item \textbf{Privacy--segmentation trade-off:} We present a systematic analysis of the privacy--segmentation trade-off over full-resolution and multiple low-resolution RGB input regimes, identifying $16\times16$ resolution as a favorable sweet point.
    \item \textbf{Real-world downstream validation:} We deploy our approach in real-world privacy-constrained scenarios, demonstrating successful downstream task execution even under severe visual degradation.
\end{itemize}
To support reproducibility and future research, our implementation is open-source\footnote{\url{https://github.com/hxy-0818/ULR2SS.git}}.

\section{Related Work}\label{sec2}
RGB-based semantic segmentation is an important capability in visual perception, but achieving semantic segmentation often conflicts with privacy requirements on visual data. 
We therefore review related work from two perspectives: privacy protection for visual perception, and semantic segmentation methods.

\subsection{Privacy Protection in Visual Perception}
As vision-based systems are increasingly deployed in privacy-sensitive and network-connected environments, protecting identifiable individuals and sensitive content has become critical. The spectrum of visual privacy protection can be broadly categorized into post-capture processing and capture-time data minimization.

\subsubsection{Post-Capture Privacy Protection}
A large body of work protects privacy after HR RGB image or videos acquisition through sanitization, redaction, anonymization or de-identification. 
At the filtering level, ~\citep{Dufaux2008tcsvt} propose region-of-interest (ROI) scrambling methods for privacy protection in video surveillance, explicitly operating on designated sensitive regions within a coded-video framework, while keeping the remainder of the frame usable.
Beyond filtering, ~\citep{Orekondy2018cvpr} formulate automatic visual redaction as a pixel-labeling problem over diverse privacy classes, and study privacy--utility trade-offs under different redaction strategies and redaction granularities.
Generative anonymization further synthesizes identity-free content~\citep{Hukkelas2019isvc}, by using a conditional generative model to produce anonymized faces conditioned on privacy-safe cues to preserve privacy. 
De-identification approaches provide a more explicit identity-removal objective. ~\citep{Wen2023iccv} perturbs 3D facial identity representations to conceal identity while preserving identity-agnostic attributes such as pose, expression, and illumination, thereby improving privacy protection without substantially compromising visual utility.
While effective in many settings, these methods still begin with information-rich visual observations, meaning that sensitive content must first be captured before protection is applied.

Another line of work aims to protect privacy via secure inference. Representative approaches include homomorphic-encryption-based inference~\citep{Gilad2016icml,Xu2024neurips}, secure multi-party computation and hybrid cryptographic protocols~\citep{Bian2020cvpr,Pang2024sp}.
However, these methods often introduce additional computation or hardware overheads, limiting practicality for real-time visual perception systems.

A complementary strategy is to replace raw imagery with more privacy-friendly representations. 
For human-centric tasks, pose skeletons and edge maps are frequently utilized~\citep{Moon2023aaai, Cai2025ipmi} to preserve sensitive information.
Similarly, SegLoc~\citep{pietrantoni2023cvpr} leverages semantic segmentation to build compact, privacy-preserving representations for visual localization. However, such structured representations are still typically derived from already acquired visual observations, mostly from information-rich RGB inputs, which can pose privacy risks of data leakage.

Taken together, post-capture methods mitigate privacy leakage after visually rich observations acquisition. As a result, they still leave a window in which privacy-sensitive content may be buffered, processed, or exposed before protection takes effect.

\subsubsection{Capture-Time Privacy Protection}
Rather than attempting to sanitize visual data after acquisition, privacy protection can also be pursued by changing the sensing modality itself so that sensitive appearance information is not captured in the first place. Depth is a representative example of such a privacy-friendlier modality~\citep{huang2025arxiv}. Recent work MOSAIC~\citep{liu2025arXiv} generates consistent, privacy-preserving scene reconstructions from multiple depth views in multi-room indoor environments, targeting scenarios in which RGB capture is restricted.

Another strategy is to minimize visual data at capture, e.g., by sensing at extreme low resolution so that identifiable details are not acquired. Kim \emph{et al.} propose an ULR camera framework for privacy-preserving robot vision with only $15 \times 15$ resolution~\citep{kim2019iros}. Ryoo et al.~\citep{Ryoo2018aaai} utilized ULR ($16 \times 12$) inputs to achieve privacy-preserving activity recognition. 
A recent user study suggests that resolution around $16 \times 16$ is privacy-preferable~\citep{huang2025arxiv}.
At such resolutions, facial regions are typically represented by only a few pixels, making reliable identity recognition difficult for both humans and machines.

Our work builds on this capture-time data-minimization by reducing privacy risks through ULR RGB acquisition, which narrows the buffering-related attack surface associated with information-rich RGB observations in connected deployments. 

\subsection{Semantic Segmentation}
Although privacy can be safeguarded through ULR capture, this typically comes at the cost of losing the spatial detail and fine-grained cues that semantic segmentation relies on (e.g., object boundaries, texture, small structures). 
To address this, we next review works on semantic segmentation.

\subsubsection{General RGB-Based Semantic Segmentation}
RGB-based semantic segmentation has emerged as a mainstream technique for visual perception, enabling dense and pixel-level understanding of visual scenes. 

CNN-based methods establish the classical end-to-end paradigm for semantic segmentation from RGB images. Fully Convolutional Networks (FCNs) ~\citep{Long2015cvpr} first demonstrated that classification backbones can be converted into fully convolutional architectures for pixel-wise prediction. Subsequent works further investigate semantic segmentation by contextual reasoning and multi-scale representation learning. 
For example, PSPNet~\citep{Zhao2017cvpr} introduces pyramid pooling to aggregate global contextual information, while DeepLabV3+~\citep{chen2018eccv} combines atrous convolution with an encoder--decoder design to improve both contextual understanding and boundary refinement. 

As semantic segmentation increasingly demands stronger long-range reasoning, Transformer-based methods have emerged as an important alternative. SETR~\citep{Zheng2021cvpr} formulates semantic segmentation as a sequence-to-sequence prediction problem using a pure Transformer encoder. SegFormer~\citep{xie2021neurips} further introduces a hierarchical Transformer encoder together with a lightweight MLP decoder, achieving strong accuracy--efficiency trade-offs across standard benchmarks. Mask2Former~\citep{Cheng2022cvpr} extends this line by unifying semantic, instance, and panoptic segmentation through masked attention and mask-based decoding.

A more recent trend scales both data and model capacity to improve transferability and generalization. Promptable and foundation-style segmentation models exemplify this direction. As a representative work, Segment Anything (SAM)~\citep{kirillov2023cvpr} introduces a promptable segmentation model trained on large-scale data and shows strong zero-shot transfer across diverse image distributions and segmentation tasks. 

Despite their strong performance and generalization, these methods are typically built on HR RGB inputs, thereby introducing privacy risks.

\subsubsection{Low-Resolution Semantic Segmentation}
In addition to standard RGB semantic segmentation, prior works have also explored recovering semantic segmentation from low-resolution RGB inputs. 

A common strategy is a two-stage pretrained pipeline, where an SR model first restores the low-resolution image and a segmentation model is then applied to the reconstructed result. For example,~\citep{caputa2024life} study this paradigm in veterinary cytology.

Beyond directly combining pretrained SR and segmentation models, some works have explored optimization strategies to improve segmentation under low-resolution inputs. In remote sensing imagery, Frizza et al.~\citep{frizza2022cviu} freeze a pretrained segmentation network as a feature extractor while training only the SR module, whereas Pereira et al.~\citep{pereira2020lagirs} investigate an alternative strategy that first upsamples the low-resolution input, for example with bicubic interpolation, and then trains only the segmentation network. Pereira et al.~\citep{pereira2020lagirs} also study a fully joint-learning setting, in which a single end-to-end model simultaneously optimizes SR and semantic segmentation objectives.

Although these approaches are closely related to our problem, they are developed either for other application domains or for low-resolution settings~(e.g. $60\times60$) that are not as extreme as the privacy-preserving ULR regime considered here. As a result, they still assume richer visual structure and more recoverable semantic cues than are available in our setting. Under our privacy-preserving ULR inputs, where semantic information is severely constrained, these methods are insufficient to deliver satisfactory segmentation performance. In contrast, our work focuses on recovering improved semantic segmentation directly from ULR RGB inputs for privacy-preserving downstream applications.

\section{Ultra-Low-Resolution Semantic Segmentation}\label{method}
\begin{figure}[!t]
  \makebox[\columnwidth][l]{%
    \includegraphics[width=1\columnwidth]{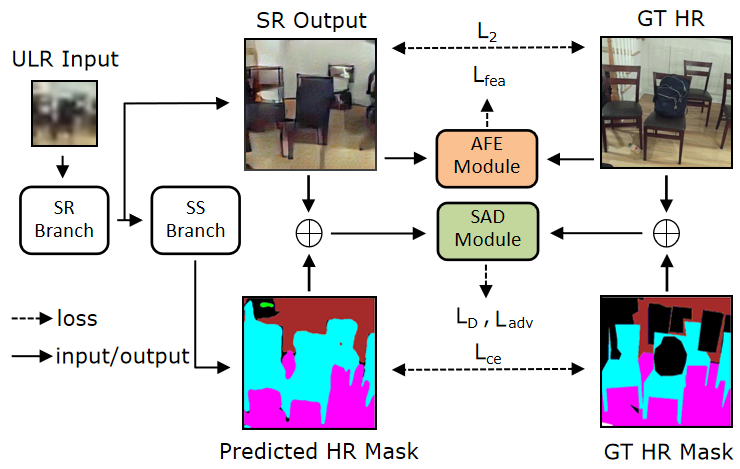}%
  }
  \caption{Overview of our proposed joint-learning framework for ultra‑low‑resolution semantic segmentation.
  Given an ULR RGB image, a GAN-based SR branch generates HR RGB, which is then fed into the SS branch (any semantic segmentation network) to predict a semantic map. 
  We integrate an AFE module to extract high-level semantic information and a SAD module to assess the realism of a concatenated super-resolved RGB image and predicted segmentation map. 
  The AFE module takes the SR output and the GT HR as input to compute the feature loss $\mathcal{L}_\mathrm{fea}$.
  The SAD module receives the concatenated pair to compute the segmentation-aware discrimination loss $\mathcal{L}_\mathrm{D}$ and the adversarial loss $\mathcal{L}_\mathrm{adv}$. 
  The SAD module is trained on the segmentation‐aware discrimination loss $\mathcal{L}_\mathrm{D}$.
  The entire network with SR and SS branches is jointly trained using the pixel-wise loss $\mathcal{L}_\mathrm{2}$, feature loss $\mathcal{L}_\mathrm{fea}$, adversarial loss $\mathcal{L}_\mathrm{adv}$ and cross-entropy loss $\mathcal{L}_\mathrm{ce}$. 
}
  \label{fig:Overview}
  \vspace{-0.5cm} 
\end{figure}

We propose a novel joint-learning framework for ULR semantic segmentation.
An overview of our framework is shown in Fig.~\ref{fig:Overview}.
Given an ULR RGB input, we first employ a Generative Adversarial Network (GAN)-based SR network to reconstruct a high‑resolution image.
This super‑resolution output is then passed through an encoder-decoder semantic segmentation network to produce a plausible segmentation map. 
We incorporate an agglomerative feature extractor~(AFE) to extract high-level semantic features from the RGB representation. 
We concatenate the predicted segmentation map with the super-resolved RGB output, the ground truth (GT) segmentation with the GT HR RGB, and design a segmentation‐aware discriminator~(SAD) to assess the realism of both the generated HR RGB image and its predicted semantic segmentation map. 
The segmentation-aware discrimination loss $\mathcal{L}_\mathrm{D}$, computed on the concatenated pair, is backpropagated separately.
Our proposed joint-learning method is trained on four losses: a pixel-wise difference between HR and super-resolved signal $\mathcal{L}_\mathrm{2}$, a feature loss $\mathcal{L}_\mathrm{fea}$, an adversarial loss on the concatenated pair $\mathcal{L}_\mathrm{adv}$, and a pixel-wise cross-entropy loss $\mathcal{L}_\mathrm{ce}$ between the predicted label and the ground-truth~(GT) label. 

\subsection{Joint-Learning Architecture} \label{S:Joint Arch}
Considering the real‐time constraints of downstream applications and the quality of super‐resolution for segmentation, we adopt the standard ESRGAN generator built upon Residual-in-Residual Dense Network (RRDN) architecture proposed by Wang~\citep{wang2018eccvworkshops}. 
Our implementation includes 23 residual-in-residual dense blocks (RRDBs), each RRDB encapsulating three dense blocks, and each dense block in turn consisting of five sub-layers (ConvBlocks).  

We employ DeepLabV3+~\citep{chen2018eccv} as the semantic segmentation module of our proposed end-to-end fully joint-learning framework. This method follows an encoder–decoder design. For the encoder, we use a pretrained ResNet-101 backbone to extract rich, multi-level feature maps, which are then passed to an Atrous Spatial Pyramid Pooling (ASPP) module to aggregate contextual information at multiple scales without reducing spatial resolution. 
The decoder upsamples the ASPP output, merges it with low-level encoder features to recover fine edges, applies a 3×3 refinement convolution, and restores full-resolution predictions through a second upsample. 

The total loss is the balanced combination of SR and semantic segmentation, with $\alpha$ controlling the influence of each loss function. 
\begin{equation}
\label{equ:L_total}
\mathcal{L}_{\mathrm{tot}} = (1 - \alpha)(\lambda_{\mathrm{1}}\mathcal{L}_{\mathrm{2}} + \lambda_{\mathrm{2}}\mathcal{L}_{\mathrm{fea}} + \lambda_{\mathrm{3}}\mathcal{L}_{\mathrm{adv}}) + \alpha\mathcal{L}_{\mathrm{ce}}
\end{equation}
For the super-generation part, we utilize a $\mathcal{L}_\mathrm{2}$ pixel-wise reconstruction term as shown in Eq.~(\ref{eq:L2}), combined with a feature loss and an adversarial loss. The scalar coefficient $\lambda$ controls the impact of each term.
\begin{equation}
\mathcal{L}_2
= \frac{1}{H\,W\,C}
\sum_{h=1}^{H}\sum_{w=1}^{W}\sum_{c=1}^{C}
\bigl(I_{gt}(h,w,c)-I_{sr}(h,w,c)\bigr)^2
  \label{eq:L2}
\end{equation}
\noindent where $H, W, C$ are the height, width and channel dimensions of the image. 
${I}_{gt}$ and ${I}_{sr}$ represent the GT and super-resolved RGB image respectively. 
The details of feature loss $\mathcal{L}_{fea}$ and extended adversarial loss $\mathcal{L}_{adv}$ are provided in Sec.~\ref{S:AFE} and Sec.~\ref{S:SAD}, respectively.
We employ the pixel-wise cross-entropy loss for semantic segmentation: 
\begin{equation}
\label{eq:L_ce}
\mathcal{L}_{\mathrm{ce}}(x, y)
= - x_{y} + \log \sum_{j} e^{x_j}
\end{equation}
where $x$ represents the raw logits and $y$ is the ground-truth class index.
The term $x_{y}$ refers to the logit corresponding to the true class, while $x_{j}$ denotes the logits for class $j$.
This loss computes the negative log-likelihood of the correct class by implicitly applying a log-softmax operation. 

\subsection{Agglomerative Feature Extractor} \label{S:AFE}
Conventional GAN‑based super‑resolution networks calculate a perceptual loss by feeding the generated and ground truth RGB images to a pretrained VGG-16/19~\citep{simonyan2015iclr}.
However, when operating on the ULR inputs, the severely degraded textures and details render VGG‑based perceptual guidance unsatisfactory. 
To better extract the semantic features for segmentation under ULR settings, we utilize the priors from a large pretrained feature extraction model.
In particular, we use \mbox{RADIOv2.5-g} version of AM‑RADIO~\citep{ranzinger2024cvpr}, which consolidates multiple large vision foundation models into an efficient student network as our feature extractor.
As the module is required only during the training phase, it does not influence the real-time requirements of robotic applications.

We propose a feature loss, which combines a $\mathcal{L}_\mathrm{1}$ loss and cosine-similarity  $\mathcal{L}_\mathrm{cos}$ term: 
\begin{equation}
\label{equ:L_feat}
\mathcal{L}_{\mathrm{fea}} = \mathcal{L}_{\mathrm{1}} \;+\; \mathcal{L}_{\mathrm{cos}}
\end{equation}
\begin{equation}
\label{eq:L1}
\mathcal{L}_{1}
= \bigl\lVert \hat{F}_\mathrm{real} - \hat{F}_\mathrm{fake} \bigr\rVert_{1}
\end{equation}
\begin{equation}
\label{eq:cos}
\mathcal{L}_{cos}
= 1 - \cos\bigl(\,\hat{F}_\mathrm{real},\,\hat{F}_\mathrm{fake})
\end{equation}
Here, $\hat{F}_\mathrm{real}$ and $\hat{F}_\mathrm{fake}$ denote the normalized features extracted for the ground-truth (real) and predicted (fake) images, respectively.
The $\mathcal{L}_\mathrm{1}$ component penalizes element-wise deviations to reinforce local detail consistency, while the cosine component aligns feature directions to improve global semantic coherence. By uniting these complementary objectives, our loss achieves more holistic feature alignment, thereby facilitating downstream semantic segmentation and improving the model's generalization ability.
$\hat{F}$ denotes the feature representation after $\mathcal{L}_\mathrm{2}$ normalization, which is used to compute the angle.

\subsection{Segmentation-Aware Discriminator} \label{S:SAD}
The key component of our approach is the Segmentation‐Aware Discriminator, which extends the conventional GAN discriminator.
A standard discriminator evaluates the visual fidelity between generated and ground‐truth RGB images and primarily encourages the generator to produce images closely resembling the GT RGB, rather than images optimized for semantic segmentation.
Under ULR conditions, where semantic information is severely limited, this exclusive focus on RGB reconstruction does not lead to improved segmentation.
Inspired by~\citep{Sushko2021iclr}, which introduces segmentation cues to the discriminator, our SAD module overcomes this limitation by jointly assessing both RGB quality and segmentation performance, thereby balancing the adversarial reconstruction loss with a segmentation-driven objective to encourage the generator to produce outputs that are beneficial for semantic segmentation. 
We introduce a segmentation-aware discrimination loss to train SAD:
\begin{equation}
\label{equ:Loss_D}
\begin{aligned}
\mathcal{L}_{\mathrm{D}}
\;=\;
\mathcal{L}_{\mathrm{BCE}}(\mathcal{D}(z_\mathrm{real}), 1)\;+\;\mathcal{L}_{\mathrm{BCE}}(\mathcal{D}(z_\mathrm{fake}), 0)
\end{aligned}
\end{equation}
\begin{equation}
\label{equ:Loss_BCE}
\mathcal{L}_{\mathrm{BCE}}(x,y) = -\Bigl[ y\log\bigl(\sigma(u)\bigr) + (1-y)\log\bigl(1-\sigma(u)\bigr) \Bigr]
\end{equation}
\begin{equation}
\label{equ:sigma}
\sigma(u) = \frac{1}{1 + e^{-u}},\quad
u \in \{\,\mathcal{D}(z_\mathrm{real}),\,\mathcal{D}(z_\mathrm{fake})\,\}
\end{equation}
\begin{equation}
\label{equ:z_define}
z_{\mathrm{real}} = \mathrm{concat}\bigl(I_{\mathrm{gt}}, S_{\mathrm{gt}}\bigr), \quad z_{\mathrm{fake}} = \mathrm{concat}\bigl(I_{\mathrm{sr}}, S_{\mathrm{pred}}\bigr)
\end{equation}
Instead of using the relativistic discrimination loss from ESRGAN, we employ the binary cross-entropy~(BCE) loss from SRGAN~\citep{ledig2017cvpr} for the discriminator, which estimates the probability that one input image comes from the true HR data distribution.
Here $\mathcal{D}(\cdot)$ outputs the discriminator logits.
$z_\mathrm{real}$ denotes the concatenated GT image $I_{\mathrm{gt}}$ with its segmentation mask~${S}_{\mathrm{gt}}$ while $z_\mathrm{fake}$ is the concatenated SR image~${I}_{\mathrm{sr}}$ with corresponding predicted semantic label ${S}_{\mathrm{pred}}$. 

The adversarial loss $\mathcal{L}_{adv}$ derived from SAD in the total loss Eq.~\ref{equ:L_total} is:
\begin{equation}
\label{equ:L_adv}
\mathcal{L}_{\mathrm{adv}}
= \mathcal{L}_{\mathrm{BCE}}\bigl(D(z_{\mathrm{fake}}),1\bigr)
\end{equation}

Moreover, we employ spectral normalization~\citep{miyato2018iclr} to scale the weights of SAD and guarantee Lipschitz-continuity.
We discover this to be essential in improving GAN training stability. By bounding the gradients of SAD, it prevents large spikes in the backpropagation signal of the generator and reduces the risk of instability.

\section{Experimental Results}

This section presents four groups of experiments. We first analyze the privacy preservation across multiple resolutions using a model-based privacy evaluator on a collected indoor privacy dataset~(Sec.~\ref{sec:ULR_threshold}). We then examine semantic segmentation performance under the ULR setting on SUN RGB-D~\citep{song2015cvpr}, including both comparisons with representative baselines and analysis of why our proposed joint-learning framework works, and further evaluate its zero-shot generalization on a real-world dataset~(Sec.~\ref{sec:ULR_results_sun}). Next, we analyze the privacy--segmentation trade-off of our method across resolutions between privacy preservation and segmentation performance~(Sec.~\ref{sec:PP_SS}). Finally, we deploy our privacy-preserving semantic segmentation method in a real-world downstream robotic object-goal navigation task under ULR setting~(Sec.~\ref{sec:real_world_exp}).

\subsection{Privacy Preservation Analysis across Resolutions} \label{sec:ULR_threshold}
Before evaluating semantic segmentation performance, it is essential to define a resolution for privacy preservation. 
Prior studies have demonstrated that ULR RGB resolutions, such as $15\times15$~\citep{kim2019iros} and $16\times12$~\citep{Ryoo2018aaai}, can effectively suppress identifiable visual details in privacy-sensitive settings. 
Consistent with this evidence, a recent user study identified $16\times16$ as a privacy-preferable resolution. 

To empirically validate its privacy preservation capability, we complement the subjective user study with an objective, model-based image privacy evaluation. 
Our goal is not to claim a formal privacy guarantee, but to compare the privacy non-leakage rate with findings from the user study.

\begin{table}[!t]
  \centering
  \renewcommand{\arraystretch}{1.2} 
  \resizebox{\columnwidth}{!}{%
  \begin{tabular}{l|cccc}
    \toprule
    Prompt & $384{\times}384$ & $32{\times}32$ & $16{\times}16$ & $8{\times}8$ \\
    \midrule
    Original & 0.0615 & 0.4231 & 0.4769 & 0.6615 \\
    Readability-Aware & 0.2692 & 0.6615 & \textbf{1.0000} & \textbf{1.0000} \\
    \bottomrule
  \end{tabular}%
  }
  \caption{Image privacy non-leakage rate comparison across four different resolutions: $384{\times}384$, $32{\times}32$, $16{\times}16$, and $8{\times}8$, evaluated under both the original and readability-aware prompts on the entire privacy dataset.}
  \label{tab:privacy_raw_rgb}
  \vspace{-0.5cm}
\end{table}

\begin{figure}[!t] 
  \centering
  \includegraphics[width=\columnwidth]{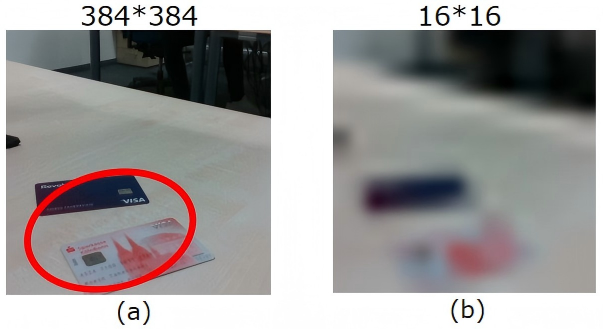}
  \caption{Example of credit card in both ($384\times384$) and ULR ($16\times16$) resolution. With the original prompt, the model flags privacy in both cases, driven by shape cues in (b) even though sensitive text is not readable.}
  \label{fig:privacy_visual}
\end{figure}

\begin{table*}[!t]
\centering
\renewcommand{\arraystretch}{1.2} 
\resizebox{\textwidth}{!}{%
\begin{tabular}{|c|c|c|c|c|} 
    \hline 
    Type & SR-Trained & Seg-Trained & Training Dataset & Description \\ 
    \hline
    
    PP & -- & \checkmark & \multirow{2.5}{*}{$384 \to 384$} & Pretrained pipeline of segmentation networks \\ 
    \cmidrule{1-3} \cmidrule{5-5} 
    
    SR+PP~\cite{caputa2024life} & $\times$ & \checkmark & & Pretrained pipeline of both SR and segmentation networks \\ 
    \hline 

    MMT-SR~\cite{frizza2022cviu} & \checkmark & $\times$ & \multirow{3.5}{*}{$16 \to 384$} & Mono-module training of SR with segmentation networks frozen \\ 
    \cmidrule{1-3} \cmidrule{5-5}
    
    MMT-Seg~\cite{pereira2020lagirs} & $\times$ & \checkmark & & Mono-module training of segmentation networks with SR network frozen\\ 
    \cmidrule{1-3} \cmidrule{5-5}
    
    FLJ~\cite{pereira2020lagirs} & \checkmark & \checkmark & & Fully joint learning of both SR and segmentation networks \\ 
    \hline 
\end{tabular}%
}
\caption{Detailed configuration of different types of pipelines.
  Each pipeline is represented by a concise type label, which distinguishes differences in module training and the training dataset resolutions. 
  $384 \rightarrow 384$ represents that both RGB input and segmentation output have resolutions of $384\times384$. 
  $16 \rightarrow 384$ represents that the RGB input is processed to $16\times16$, but the segmentation output is predicted at resolution $384\times384$.}
\label{tab:config}
\vspace{-0.5cm}
\end{table*}

\textbf{Dataset and Protocol:} Since SUN RGB-D~\citep{song2015cvpr} is a public indoor-scene dataset and does not provide annotations for privacy-sensitive content, it cannot directly support privacy-leakage evaluation. 
While prior studies have investigated image privacy assessment, most rely on dataset-specific privacy taxonomies and thus require predefined sensitive classes.
We therefore follow the protocol of Samson et al.~\citep{samson2024arxiv}, which was validated on the PrivBench benchmark. We constructed a targeted indoor privacy dataset comprising 130 images across four sub sensitive categories: face, debit card, passport, and private chat. 
Each image contains one privacy category. 

\textbf{Evaluation Metric:} The model~\citep{samson2024arxiv} outputs a \textit{privacy rate} indicating the presence of privacy-related content within an image. 
From this, we derive a complementary privacy non-leakage metric, \text{P-non}, computed as $1 - \textit{privacy rate}$, which serves as a model-based proxy for privacy preservation.
A larger \text{P-non} therefore indicates lower privacy leakage through model-based recognition.

Building on~\citep{samson2024arxiv}, we assess privacy preservation for raw RGB inputs across four resolutions: $384{\times}384$, $32{\times}32$, $16{\times}16$, and $8{\times}8$. 
We first evaluate privacy preservation under the original prompt, \emph{"Analyze the image provided. Does the attached image contain private information? Answer only with yes or no."}
As shown in the Table~\ref{tab:privacy_raw_rgb}, \text{P-non} naturally increases as the image resolution decreases. 
However, presence-only prompting can be misleading at ULR. As illustrated in Fig.~\ref{fig:privacy_visual}, Fig.~\ref{fig:privacy_visual}(a) shows a $384 \times 384$ image where a credit card is highlighted (red circle), while Fig.~\ref{fig:privacy_visual}(b) shows its $16 \times 16$ counterpart. Even though the private information in Fig.~\ref{fig:privacy_visual}(b) is not readable, the model answers “yes” under the original prompt by relying on coarse shape cues (e.g., the card outline).

To mitigate this issue, we therefore introduce a readability-aware prompt: \emph{"Analyze the image provided. Does the attached image contain private information and if yes, is the private information readable? Answer only with yes or no."}
Under the readability-aware prompt, ULR images ($\leq$ $16 \times 16$) achieve $\text{P-non = 1.0000}$ across the entire dataset, while $32 \times 32$ remains lower, as shown in the Table~\ref{tab:privacy_raw_rgb}. 
This reveals a clear transition around $16 \times 16$, which is consistent with the recent user-study finding that $16 \times 16$ is a promising privacy-preserving ULR.
Detailed results on privacy preservation of each sensitive category are provided in the Appendix~\ref{sec:p_types}.

Taking all three evidence, we therefore take $16 \times 16$ as ULR RGB inputs to recover semantic segmentation in this work. 

\subsection{Ultra-Low-Resolution Semantic Segmentation} \label{sec:ULR_results_sun}
Having determined $16 \times 16$ as the privacy-preserving ULR setting, we next evaluate semantic segmentation under this regime. In this section, we first compare our method with diverse representative baselines on \mbox{SUN RGB-D} in terms of semantic segmentation performance and inference time. We then analyze why our proposed joint-learning design is effective under extreme visual degradation by examining the relationship between super-resolution and ULR semantic segmentation. Finally, we assess the zero-shot generalization ability of our model on a collected real-world data.

\subsubsection{Quantitative Evaluation on Standard Benchmark} \label{sec:benchmark_results}
We assume the ULR semantic segmentation workflow is consistent across all methods, taking a single RGB image and producing a semantic segmentation map with a fixed output size of $384 \times 384$.

\textbf{Dataset:} We trained our network on the SUN RGB-D dataset~\citep{song2015cvpr}, a widely used benchmark for indoor semantic segmentation. This dataset contains images from NYU Depth v2~\citep{silberman2012eccv}, Berkeley B3DO~\citep{Janoch2011iccvworkshop}, and SUN3D~\citep{Xiao2013iccv}, comprising a total of 10,335 images captured across diverse indoor environments.
Each image was cropped to $384 \times 384$ pixels to serve as our high‐resolution ground truth. We randomly split the dataset into a training set~(9,000~images), a validation set~(668~images), and a test set~(667~images). To better train different methods, we processed the dataset into two groups of resolutions: $384 \rightarrow 384$ and $16 \rightarrow 384$.
$384 \rightarrow 384$ specifies that both RGB input and the corresponding segmentation mask have the same resolution of $384\times384$.
$16 \rightarrow 384$ indicates that the RGB input is resized to $16\times16$, but the segmentation output is predicted at resolution $384\times384$.

\textbf{Baselines:} To assess the segmentation performance of our approach, we compare our proposed method against five categories of setup-consistent pipelines with different SR and segmentation networks as shown in Table~\ref{tab:config}.
For the SR module, we adopt ESRGAN~\citep{wang2018eccvworkshops}. 
For the segmentation module, we consider three architectures: DeepLabv3+~\citep{chen2018eccv} and DFormer~\citep{wang2023arXiv}, as well as a SAM-based variant built upon \citep{kirillov2023cvpr}. 
Since SAM~\citep{kirillov2023cvpr} is originally designed for promptable, class-agnostic mask prediction rather than multi-class semantic segmentation, we follow the adapter-based adaptation strategy of Medical SAM Adapter~\citep{wu2025mia} to equip SAM with a semantic segmentation head, and finetune it on SUN RGB-D. 
Combining five pipeline variants with three segmentation backbones results in $5\times3=15$ baselines in total.

\textbf{Evaluation Metrics:} Following~\citep{chen2018eccv}, we use mean Intersection-over-Union (mIoU) and mean class accuracy (mAcc) to quantify the semantic segmentation performance. 

\textbf{Training Setup:}
Training was conducted on two NVIDIA A100 GPUs with a batch size of 16, and we evaluated our approach on a laptop equipped with a NVIDIA GeForce RTX 4060. 
We set the training hyperparameters as follows: $\lambda_{1}$ = 0.5, $\lambda_{2}$ = 0.01, $\lambda_{3}$ = 0.01 and $\alpha$ = 0.3. 
Detailed ablation results on the selection of $\alpha$ for mitigating the objective conflict between the super-resolution and semantic segmentation are provided in Appendix~\ref{sec:alpha}.

We first pretrained the SR model using the original ESRGAN~\citep{wang2018eccvworkshops} losses: a pixel-wise MAE loss, a vanilla adversarial loss, and a VGG-based perceptual loss. This strategy accelerated convergence and mitigated the early‐stage instabilities. Next, we integrated the DeepLabv3+ segmentation network~\citep{chen2018eccv} and jointly trained both modules. 
Both stages used the same learning rate of 1e-4 and employed ADAM optimizer with $\beta_1$ = 0.9 and $\beta_2$ = 0.999. We trained the model for 100~epochs and selected the best model according to the highest mIoU on the validation set.

To ensure a controlled and fair comparison that isolates the effect of our proposed modules, we disable all data augmentation across all methods. 
This follows a unified training scheme and prevent performance variations arising from method-specific augmentation, allowing improvements to be attributed to the proposed architecture.

\begin{table}[!t]
  \centering
  \renewcommand{\arraystretch}{1.2}
  \setlength{\tabcolsep}{6pt}
  \resizebox{\columnwidth}{!}{%
    \begin{tabular}{|c|c|c|c|c|c|}
      \hline
      Input & Type & SS-Network & mIoU$\uparrow$ & mAcc$\uparrow$ & Time\,(s)$\downarrow$\\ 
      \hline

      \multirow{3}{*}{384$\times$384}
        & \multirow{3}{*}{PP}    & SAM  & \textbf{0.4201} & \textbf{0.5426} & 0.0821 \\ \cmidrule{3-6}
        &                        & DF   & 0.4103 & 0.5352 & 0.2201 \\ \cmidrule{3-6}
        &                        & Deep & 0.4194 & 0.5369 & \textbf{0.0075} \\ 
      \hline

      \multirow{18}{*}{16$\times$16}
        & \multirow{3}{*}{PP}    & SAM  & 0.0359 & 0.0468 & 0.0800 \\ \cmidrule{3-6}
        &                        & DF   & 0.0560 & 0.1878 & 0.2178 \\ \cmidrule{3-6}
        &                        & Deep & 0.0716 & 0.1478 & \textbf{0.0075} \\ \cmidrule{2-6}
        
        & \multirow{3}{*}{SR+PP} & SAM  & 0.0757 & 0.0935 & 0.1263 \\ \cmidrule{3-6}
        &                        & DF   & 0.1387 & 0.2157 & 0.2791 \\ \cmidrule{3-6}
        &                        & Deep & 0.1241 & 0.1703 & 0.0542 \\ \cmidrule{2-6}
        
        & \multirow{3}{*}{MMT-SR}& SAM  & 0.1450 & 0.2073 & 0.1305 \\ \cmidrule{3-6}
        &                        & DF   & 0.1311 & 0.1885 & 0.2873 \\ \cmidrule{3-6}
        &                        & Deep & 0.1640 & 0.2348 & 0.0577 \\ \cmidrule{2-6}
    
        & \multirow{3}{*}{MMT-Seg}& SAM & 0.2736 & 0.3835 & 0.1298 \\ \cmidrule{3-6}
        &                        & DF   & 0.2855 & 0.3901 & 0.2699 \\ \cmidrule{3-6}
        &                        & Deep & 0.2580 & 0.3480 & 0.0576 \\ \cmidrule{2-6}
        
        & \multirow{6}{*}{FJL}   & SAM           & 0.2317 & 0.3315 & 0.1335 \\ \cmidrule{3-6}
        &                        & DF            & 0.2568 & 0.3603 & 0.2878 \\ \cmidrule{3-6}
        &                        & Deep          & 0.2575 & 0.3601 & 0.0559 \\ \cmidrule{3-6}
        &                        & Ours w/o SAD  & 0.2824 & 0.4208 & 0.0577 \\ \cmidrule{3-6}
        &                        & Ours w/o AFE  & 0.3021 & 0.4503 & 0.0525 \\ \cmidrule{3-6}
        &                        & \textbf{Ours} & \textbf{0.3101} & \textbf{0.4620} & 0.0567 \\ 
      \hline
    \end{tabular}%
  }
  \caption{Semantic segmentation and inference time results comparison across different baselines and our proposed method on \mbox{SUN RGB-D} test dataset. As can be seen, our method attains the best mIoU and mAcc among all baselines with input 16$\times$16 while maintaining a real-time inference.}
  \label{tab:ulr_ss_results}
  \vspace{-0.5cm}
\end{table}

\begin{table*}[!t]
  \centering
  \resizebox{0.9\textwidth}{!}{%
    \begin{tabular}{l|cccc|ccc|cc}
      \toprule
      \multirow{2}{*}{Method} 
    & \multicolumn{4}{c}{RGB Reconstruction Fidelity} 
    & \multicolumn{3}{c}{Segmentation Accuracy} 
    & \multicolumn{2}{c}{Semantic Accuracy} \\
    \cmidrule(lr){2-5} \cmidrule(lr){6-8} \cmidrule(lr){9-10}
    & PSNR$\uparrow$  & SSIM$\uparrow$  & LPIPS$\downarrow$  
    & FID$\downarrow$ & ARI$\uparrow$ & Covering$\uparrow$  
    & BF$\uparrow$  & mIoU$\uparrow$  & mAcc$\uparrow$ \\
      \midrule
      PP-Deep         & \textbf{20.03}  & \textbf{0.6124} & 0.7088  & 263.78 & 0.1220 & 0.3666     & 0.0514 & 0.0716 & 0.1478\\
      SR+PP-DF         & 20.02  & 0.5598 & 0.5631  & 218.27 & 0.3370 & 0.4928     & 0.1175 & 0.1387 & 0.2157\\
      MMT-SR-Deep       & 15.91  & 0.4382 & 0.5151  & 104.43 & 0.2885 & 0.4540     & 0.1303 & 0.1640 & 0.2348\\
      MMT-Seg-DF           & 20.02  & 0.5598 & 0.5631  & 218.27 & 0.3704 & 0.5000     & 0.1377 & 0.2855 & 0.3901 \\
      FJL-Deep        & 12.22  & 0.3218 & 0.5839  & \textbf{100.94} & 0.3524 & 0.4917     & \textbf{0.1452} & 0.2575 & 0.3601 \\
      \textbf{Ours}         & 15.07  & 0.4059 & \textbf{0.4892}  & 242.73 & \textbf{0.4010} & \textbf{0.5271}     & 0.1359 & \textbf{0.3101} & \textbf{0.4620} \\
      \bottomrule
    \end{tabular}%
  }
  \captionof{table}{Quantitative comparison of super‐resolution outputs, segmentation and semantic segmentation results on \mbox{SUN RGB-D} test dataset. Given the ultra‐low‐resolution $16\times16$ inputs, we compare the best five baselines from each category and our proposed approach. 
  As shown in the table, our method achieves the best performance in semantic segmentation (mIoU \& mAcc) as well as segmentation ARI and Covering, despite not obtaining the strongest RGB reconstruction metrics. 
  This indicates that segmentation performance is not directly positively correlated with RGB image quality.
  }
  \label{tab:quantitative metrics results}
  \vspace{-0.3cm} 
\end{table*}

\begin{figure*}[!t]
  \centering
  \includegraphics[width=1.0\textwidth]{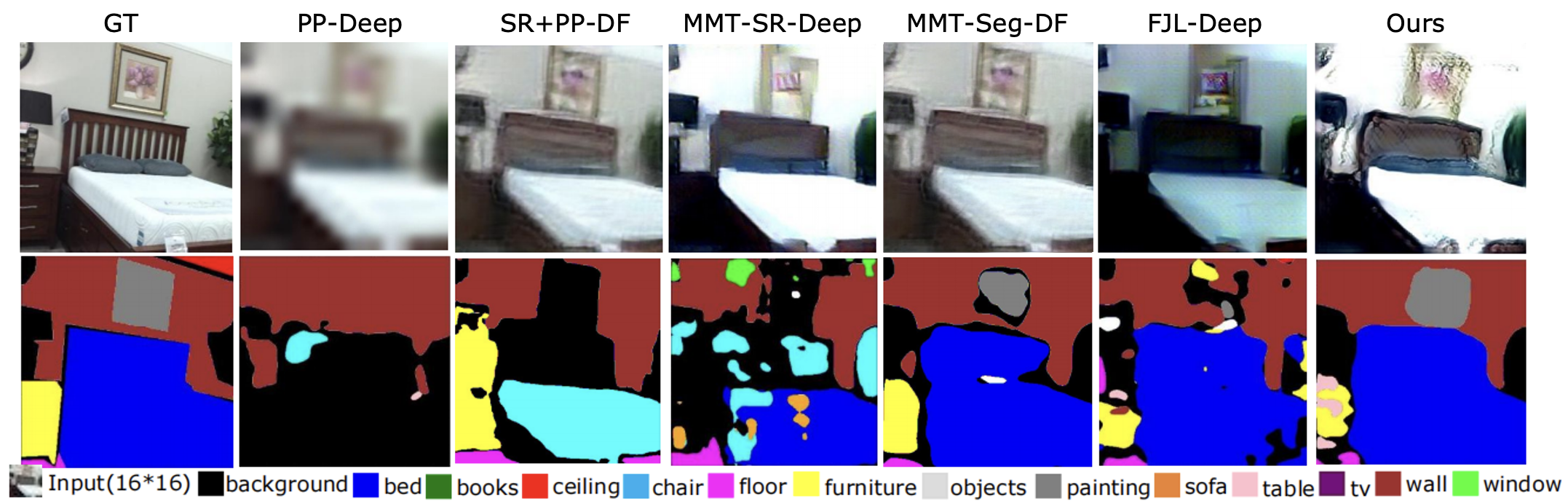}
  \caption{
  Visual comparison of super‐resolution RGB images and semantic segmentation maps. We compare the visualization results of super‐resolved outputs and predicted semantic‐segmentation produced by the best five baselines and our proposed method against the GT, under the setting of $16 \times 16$ input.
  As shown from the visualization result, our method predicts the most accurate semantic segmentation mask compared to all baselines. 
  Our method amplifies object‐boundary contrast to favor region delineation and achieves the best semantic segmentation performance at the expense of boundary segmentation.
  }
  \label{fig:visual}
  \vspace{-0.3cm} 
\end{figure*}

\textbf{Quantitative Evaluation Results}: We compare our method against different baselines on \mbox{SUN RGB-D} as shown in Table~\ref{tab:ulr_ss_results}.

\textit{Upper Bound at Full Resolution.}
We first establish an upper bound by evaluating three segmentation networks on full-resolution ($384 \times 384$) inputs.
Among them, PP-SAM achieves the highest mIoU and mAcc, providing a practical upper limit for ULR segmentation performance.

\textit{Effect of Ultra-Low Resolution.}
At an ULR of $16 \times 16$, directly applying segmentation networks pretrained on full-resolution data (PP) causes a drastic drop in both mIoU and mAcc across all three architectures, revealing the strong resolution sensitivity of this task.

\textit{SR and Mono-Module Training.}
Incorporating a SR module prior to segmentation (SR+PP) brings only marginal improvements, indicating that conventional SR restoration contributes limited semantic benefit.
When the segmentation network is frozen and only the SR module is trained (MMT-SR), performance gains remain minimal, suggesting that the SR alone cannot recover high-level semantics.
Conversely, freezing SR while fine-tuning the segmentation network (MMT-Seg) yields noticeable improvements, though performance still falls far below the full-resolution upper bound.

\textit{Joint Learning.}
A straightforward joint learning strategy (FJL), which simultaneously trains SR and segmentation branches, unexpectedly degrades accuracy, performing even worse than MMT-Seg.
This result suggests that naive coupling of the two modules leads to optimization conflicts and suboptimal feature alignment.

\textit{Our Method and Ablation Study.}
All baselines struggle under $16 \times 16$ input resolution, highlighting the need for a more effective framework.
Our proposed method achieves the highest mIoU and mAcc among all baselines, demonstrating that our SAD and AFE effectively enhance semantic recovery within the FJL-style architecture.
Removing either component noticeably degrades both mIoU and mAcc, while removing both reduces the system to the baseline FJL-Deep configuration and leads to the largest performance drop, confirming the necessity of both modules.

\textit{Inference Efficiency.}
We compare forward inference time on a NVIDIA GeForce RTX 4060 GPU.
Our approach processes each image in $0.0567\,s$ (about 17.6 FPS), meeting the real-time requirements of downstream applications.

\begin{table*}[!t]
  \centering
  \renewcommand{\arraystretch}{1.2}
  \resizebox{0.65\textwidth}{!}{%
    \begin{tabular}{|c|c|c|c|c|c|c||c|}
      \hline
      Method & \makecell{PP\\-Deep} & \makecell{SR+PP\\-DF} & \makecell{MMT-SR\\-Deep} & \makecell{MMT-Seg\\-DF} & \makecell{FJL\\-Deep} & \textbf{Ours} & \makecell{GT\\RGB} \\ \hline
      MPICD & 0.2429 & 0.2284 & \textbf{0.3429} & 0.2284 & 0.2765 & 0.3248  & 0.2524 \\ \hline
    \end{tabular}%
  }
  \caption{Mean Pairwise Inter-Class Color Distance calculated on the SR outputs of each method on SUN RGB-D test set.}
  \label{tab:mpicd}
  \vspace{-0.6cm}
\end{table*}

\subsubsection{Analysis Between Super-Resolution and Ultra-Low-Resolution Semantic Segmentation} \label{sec:SR_SS}

To better understand why naive FJL framework degrades semantic segmentation relative to MMT-Seg under ULR inputs, and why our proposed design improves upon it, we analyze the intermediate super-resolved images together with their corresponding segmentation outputs in Table~\ref{tab:quantitative metrics results} and Fig.~\ref{fig:visual}.

Our proposed pipeline outputs only semantic segmentation maps, without generating or storing intermediate super-resolved images. The intermediate super-resolved results shown here in Fig.~\ref{fig:visual} are solely for visualization purposes to facilitate analysis of the relationship between super-resolution and ULR-based semantic segmentation. The generated SR do not raise privacy concerns (Further analysis regarding the image privacy can be referred to Sec. ~\ref{sec:PP_SS}).

\textbf{Dataset:} To ensure a consistent comparison, all intermediate SR outputs and predicted semantic masks analyzed in this section are extracted from the \mbox{SUN RGB-D} test set. 

\textbf{Evaluation Metrics:} To assess the RGB reconstruction fidelity, following~\citep{kong2023rs}, we employ four widely used image-quality metrics: Peak Signal‐to‐Noise Ratio (PSNR), Structural Similarity Index Measure (SSIM), Learned Perceptual Image Patch Similarity (LPIPS) and Fréchet Inception Distance (FID). For the evaluation of segmentation performance, we report two region‐based indices, Adjusted Rand Index (ARI), Segmentation Covering (Covering), and one boundary‐based metric, Boundary F‐measure (BF)~\citep{arbelaez2010tpami}. For semantic accuracy, we report two overall semantic segmentation metrics (mIoU and mAcc).

\begin{table*}[!t]
  \centering
  \renewcommand{\arraystretch}{1.2}
  \resizebox{0.75\textwidth}{!}{%
    \begin{tabular}{|c|c|c|c|c|c|c|}
      \hline
      Method & \makecell{PP-SAM} & \makecell{SR+PP-DF} & \makecell{MMT-SR-Deep} & \makecell{MMT-Seg-DF} & \makecell{FJL-Deep} & \textbf{Ours} \\ \hline
      mIoU$\uparrow$   & 0.0456  & 0.1113   & 0.1650      & 0.2622       & 0.2391   & \textbf{0.3177} \\ \hline
      mAcc$\uparrow$   & 0.0792  & 0.1885   & 0.2699      & 0.3873       & 0.3648   & \textbf{0.4608} \\ \hline
    \end{tabular}%
  }
  \caption{Semantic segmentation test on $16\times16$ images of our collected real-world dataset. 
  Our method achieves the best zero-shot domain generalization semantic segmentation performance on both mIoU and mAcc among all baselines.
  }
  \label{tab:test16 real world}
  \vspace{-0.6cm}
\end{table*}

\textbf{The Disconnect Between RGB Reconstruction and Semantic Segmentation:} Table~\ref{tab:quantitative metrics results} reveals that generic super-resolution quality does not reliably predict downstream semantic segmentation performance under extreme ULR conditions:
\begin{itemize}
    \item Under ULR conditions, higher PSNR and SSIM do not necessarily lead to better semantic segmentation performance. 
    \item Under ULR conditions, lower LPIPS and FID do not guarantee better semantic segmentation performance.
\end{itemize}

PP-Deep achieves the highest PSNR \& SSIM, while exhibits the worst semantic segmentation performance. Likewise, MMT-Seg-DF attains higher PSNR \& SSIM than our approach, yet it does not outperform our method in semantic segmentation. These results indicate that, under extreme ULR inputs, optimizing for pixel fidelity to encourage visually smoother, distortion-minimizing reconstructions, does not guarantee preservation or recovery of the class-discriminative structures required for semantic labeling. 

A similar disconnect appears for perceptual and distributional image quality. MMT-SR-Deep achieves a lower LPIPS score than MMT-Seg-DF, yet yields inferior segmentation performance. FJL-Deep attains the best FID among the compared baselines, suggesting that its generated outputs are distributionally closest to natural images, but its semantic segmentation remains worse than both MMT-Seg-DF and our method. Thus, better perceptual realism or stronger alignment with the distribution of natural RGB images does not directly translate into better ULR semantic segmentation.

This behavior is consistent with the perception--distortion trade-off~\citep{Blau2018cvpr}. Under severe ill-posedness at 16$\times$16 resolution, pushing the SR branch toward perceptual realism tends to introduce visually plausible high-frequency details that are not guaranteed to be spatially or semantically aligned with the ground truth. When restoration and segmentation are naively optimized jointly using shared representations, these hallucinated variations can interfere with the segmentation objective. This explains why FJL-Deep, despite strong perceptual quality, fails to improve semantic segmentation. In contrast, MMT-Seg-DF freezes the SR branch and finetunes only the segmentation network, avoiding additional interference from the restoration objective. Its stronger performance relative to FJL-Deep further shows that naive joint optimization is not sufficient for effective ULR semantic segmentation.

\textbf{Why Our Proposed FJL Architecture Improves:} 
Motivated by the perception–distortion trade-off, we do not aim to maximize generic reconstruction fidelity or perceptual realism.
Our method intentionally relaxes RGB reconstruction fidelity to favor semantic consistency, achieving the highest region-level accuracy and overall semantic segmentation scores. 
As shown in Table~\ref{tab:quantitative metrics results}, our method achieves the best region-level segmentation metrics and the strongest overall semantic segmentation performance. 
Qualitatively as shown in Fig.~\ref{fig:visual}, although our reconstructed RGB images do not aim to match the ground-truth appearance pixel by pixel, they exhibit clearer inter-object transitions and more semantically useful structure for subsequent labeling.
The slightly lower BF score may result from color bleeding near object edges. 

To further analyze why our SR outputs benefit ULR segmentation, we measure Mean Pairwise Inter-Class Color Distance (MPICD).
A higher MPICD indicates greater chromatic separability between classes.
As shown in Table~\ref{tab:mpicd}, our method achieves higher MPICD than the ground truth, suggesting stronger inter-class contrast.
However, excessive separability alone does not ensure accuracy: MMT-SR-Deep attains the largest MPICD but yet still produces poor semantic segmentation when the segmentation module is frozen.
Feeding its SR outputs into our segmentation network improves its performance, with mIoU value increasing from 0.1640 to 0.2463, yet still below our method~(0.3101). These observations highlight that effective ULR segmentation requires not only contrast enhancement in the restored image space, but also joint SR–segmentation optimization that aligns restoration with semantic objectives, rather than naïvely coupling objectives that may drift toward perceptual realism.

\subsubsection{Zero-shot Domain Generalization} 
To further demonstrate the robustness of our proposed method and align with the real-world robot experiments presented later, we next evaluate its zero-shot domain generalization capability.

\textbf{Dataset and Protocol:} We conduct this evaluation using a custom real-world indoor dataset comprising 90~annotated RGB and semantic segmentation pairs, following the data collection scheme of the NYU Depth V2~\citep{silberman2012eccv} dataset. Specifically, we directly evaluated our model on these newly collected images without any domain-specific fine-tuning. This allows us to evaluate the applicability and generalization of our method.

\textbf{Evaluation Results:} Table~\ref{tab:test16 real world} shows the semantic segmentation results on our collected indoor-scene dataset under the ULR setting.
Our method achieves the highest mIoU and mAcc among the best five baselines, representing a substantial improvement over the strongest baseline (MMT-Seg-DF).
This result closely matches our performance on the \mbox{SUN RGB-D} test set, demonstrating the robustness of our approach and its strong zero-shot domain generalization ability at ULR conditions.

\begin{table}[!t]
  \centering
  \renewcommand{\arraystretch}{1.2} 
  \resizebox{\columnwidth}{!}{ 
    \begin{tabular}{l|ccc}
      \toprule
      Prompt 
      & \shortstack[c]{$32{\times}32$\\$\to 384{\times}384$}
      & \shortstack[c]{$16{\times}16$\\$\to 384{\times}384$}
      & \shortstack[c]{$8{\times}8$\\$\to 384{\times}384$} \\
      \midrule
      Original & 0.2185 & 0.6114 & 0.6822 \\
      Readability-Aware & 0.4876 & \textbf{1.0000} & \textbf{1.0000} \\
      \bottomrule
    \end{tabular}
  } 
  \caption{Image privacy non-leakage rate comparison of $384{\times}384$ super-resolved RGB images generated from $32{\times}32$, $16{\times}16$ and $8 \times 8$, evaluated under both the original and readability-aware prompts on the entire privacy dataset.
  Each entry reports the $\text{P-non}$ score.}
  \label{tab:privacy_generated_rgb}
\end{table}

\subsection{Privacy--Segmentation Trade-Off Analysis} \label{sec:PP_SS}
Having established $16 \times 16$ as a privacy-preserving ULR resolution and evaluated semantic segmentation under this setting, we now jointly analyze privacy preservation and semantic segmentation performance across resolutions. 

\textbf{Evaluation Setup:} The privacy analysis in this section follows the same indoor privacy dataset, evaluation protocol, and privacy metric as those introduced in Sec.~\ref{sec:ULR_threshold}. For semantic segmentation, we adopt the same evaluation criteria as in Sec.~\ref{sec:benchmark_results}, namely mIoU and mAcc.

\subsubsection{Privacy Preservation Analysis in Generated RGBs} 
Before examining the privacy--segmentation trade-off, we first evaluate the privacy preservation of super-resolved RGB reconstructions. 
As shown from Table~\ref{tab:privacy_generated_rgb}, under the original prompt, the P-non gap from $32\times32$ to $16\times16$ is substantial, whereas the gain from $16\times16$ to $8\times8$ is comparatively small. This suggests that $16\times16$ already lies within a practically privacy-preserving regime, while further reducing the resolution to $8\times8$ yields only marginal additional benefit under the current evaluator. 

A similar transition is also reflected under our readability-aware prompt. For $32 \times 32$ resolution, the generated $384 \times 384$ SR images obtain low P-non values, indicating that the generated outputs still contain private content that is considered readable by the evaluator.
In contrast, when applying SR to $16 \times 16$ and $8 \times 8$ images, the $\text{P-non}$ scores remain at 1.0000 under the readability-aware prompt, indicating that SR does not reintroduce readable private information under ULR setting.

\begin{table}[!t]
  \centering
  \renewcommand{\arraystretch}{1.2}
  \resizebox{0.8\columnwidth}{!}{%
    \begin{tabular}{|c|c|c||c|}
      \hline
      Method & mIoU & mAcc & P-non \\ \hline
      PP-SAM-384 & \textbf{0.3555} & \textbf{0.4597} & 0.2692 \\ \hline
      Ours-32    & 0.2979 & 0.4193 & 0.6615 \\ \hline
      Ours-16    & 0.2719 & 0.3970 & \textbf{1.0000} \\ \hline
      Ours-8     & 0.1617 & 0.2921 & \textbf{1.0000} \\ \hline
    \end{tabular}%
  }
  \caption{Analysis between privacy preservation and semantic segmentation performance under our readability-aware prompt. ``-384'', ``-32'', ``-16'' and ``-8'' indicate the input resolutions.}
  \label{tab:trade_off}
  \vspace{-0.6cm}
\end{table}

\begin{figure*}[!t]
  \centering
  \includegraphics[width=1.0\textwidth]{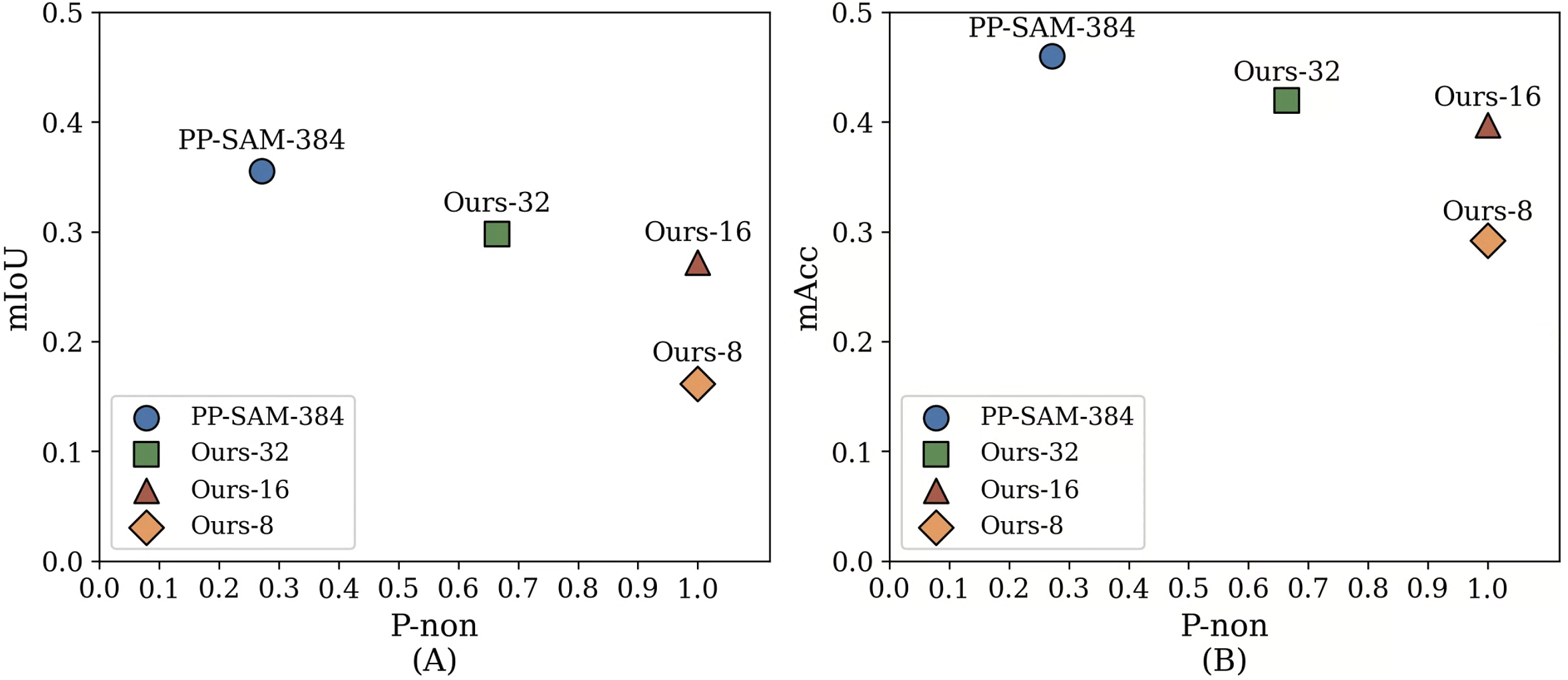}
  \caption{Trade-off between privacy non-leakage and semantic segmentation performance~(measured by mIoU (A) and mAcc (B)) across resolutions. As can be seen, $16\times16$~(Ours-16) provides the most favorable trade-off between privacy preservation and segmentation performance.
  }
  \label{fig:tradeoff}
  \vspace{-0.3cm} 
\end{figure*}

\subsubsection{Privacy-Segmentation Trade-Off Across Resolutions} 
Using the readability-aware prompt, we further analyze the trade-off between privacy preservation and semantic segmentation performance of our method across resolutions. As the full-resolution reference, we select PP-SAM-384, which attains the highest mIoU and mAcc among full-resolution PP methods. 
For a fair comparison, our resolution variants (Ours-32/16/8) are fine-tuned on \mbox{SUN RGB-D} at their respective input resolutions using the same training protocol as Ours-16, and all mIoU/mAcc results reported in Table~\ref{tab:trade_off} are obtained by directly performing zero-shot inference on the Privacy dataset as introduced in Sec~\ref{sec:ULR_threshold} without additional adaptation. Fig.~\ref{fig:tradeoff} visualizes the trade-off between privacy non-leakage rate and segmentation performance using mIoU~(A) and mAcc~(B), respectively.

We adopt Pareto optimization analysis, where a method is Pareto-optimal if no other method improves both privacy (higher P-non) and segmentation performance (higher mIoU/mAcc) simultaneously.  
Table~\ref{tab:trade_off} and Fig.~\ref{fig:tradeoff}~(A)--(B) consistently show a clear privacy--segmentation trade-off across resolutions. PP-SAM-384 yields the highest mIoU and mAcc, but also the lowest privacy non-leakage rate. 
In terms of our proposed method, Ours-32 and Ours-16 deliver similar semantic segmentation results, yet Ours-32 falls short in privacy preservation ability. 
Furthermore, although Ours-8 matches the privacy non-leakage rate as of Ours-16, Ours-8 suffers from a severe degradation in semantic segmentation performance.
Therefore, the Pareto frontier analysis suggests that the \textbf{$16\times16$} setting (Ours-16) emerges as the sweet point, representing a favorable trade-off between privacy preservation and segmentation performance.

\subsection{Privacy-Preserving Semantic Object-Goal Navigation} \label{sec:real_world_exp}
Having identified $16\times16$ as a favorable privacy--segmentation sweet point, we next deploy our privacy-preserving semantic segmentation method in a real-world robotic object-goal navigation task under ULR setting.

\textbf{System Overview:} We implement an end-to-end privacy-preserving semantic object-goal navigation system comprising three modules. 
Given a target object, the robot first starts the \textbf{SemanticRecovery} module.
If the target object is found, the robot switches to the \textbf{ObjectGoalNavigation} module to approach it.
If the target object is not detected, our system calls the \textbf{FloorBasedNavigation} module to determine the next best waypoint for exploration.
We randomly sample midpoints at fixed intervals to define the robot’s waypoints.
Foreground pixels deviating horizontally from fitted boundary lines beyond a threshold are identified as branch points, indicating potential new area entrances.
If no further waypoints are available, our system concludes that the target object is not found.
Otherwise, the robot moves to the new waypoint, and the loop continues.
Further details of our proposed navigation system are provided in Appendix~\ref{sec:nav_system}.

\textbf{Experimental Setup:} Fig.~\ref{fig:teaser} shows the application scenario of our method.
The experiments were conducted using the OrionStar GreetingBot Mini\footnote{\url{https://en.orionstar.com/mini.html}} mobile robot, with only the onboard RGB camera activated for perception.
ULR images are simulated by downsampling full-resolution images to $16 \times 16$ using bicubic interpolation.
We conducted experiments using four target objects: sofa, chair, table, and painting.
For each object, the robot is initialized from two starting locations: one in the corridor and the other inside the room.
Each object-starting point pair is tested five times, resulting in a total of 40 trials.

\textbf{Baselines and Evaluation Metric:} We compare our proposed method against the top-performing baselines from MMT-Seg and FJL categories, respectively on SUN RGB-D and our collected dataset from the real world.
The success rate is used for evaluation and computed as the proportion of trials in which the robot successfully navigates from the starting location to the target object.
A success trial is defined when the segmented pixels of the target object occupy more than 40\% of the image.

\begin{table}[!t]
  \centering
  \renewcommand{\arraystretch}{1.2}
  \resizebox{0.8\columnwidth}{!}{%
    \begin{tabular}{|c|c|}
      \hline
      Method & Success Rate\,(\%)$\uparrow$ \\ \hline
      FJL-Deep & 62.5 \\ \hline
      MMT-Seg-DF & 65.0 \\ \hline
      \textbf{Ours} & 85.0 \\ \hline 
      PP-SAM-384 (Full) & \textbf{92.5} \\ \hline
    \end{tabular}%
  }
  \caption{Comparison of privacy-preserving semantic object-goal navigation success rate.
  We conducted a total of 40 trials, testing four target objects from two distinct starting locations. 
  Our proposed method achieves the highest navigation success rate, surpassing FJL-Deep and MMT-Seg-DF, and remains competitive with the full-resolution upper bound PP-SAM-384. 
  Thus, our improved ULR semantic segmentation results increase the success rate of object-goal navigation in real-world scenarios with privacy constraints.}
  \label{table:success_rate}
  \vspace{-0.3cm}
\end{table}

\begin{figure}[!t]
  \makebox[\columnwidth][r]{%
    \includegraphics[width=1\columnwidth]{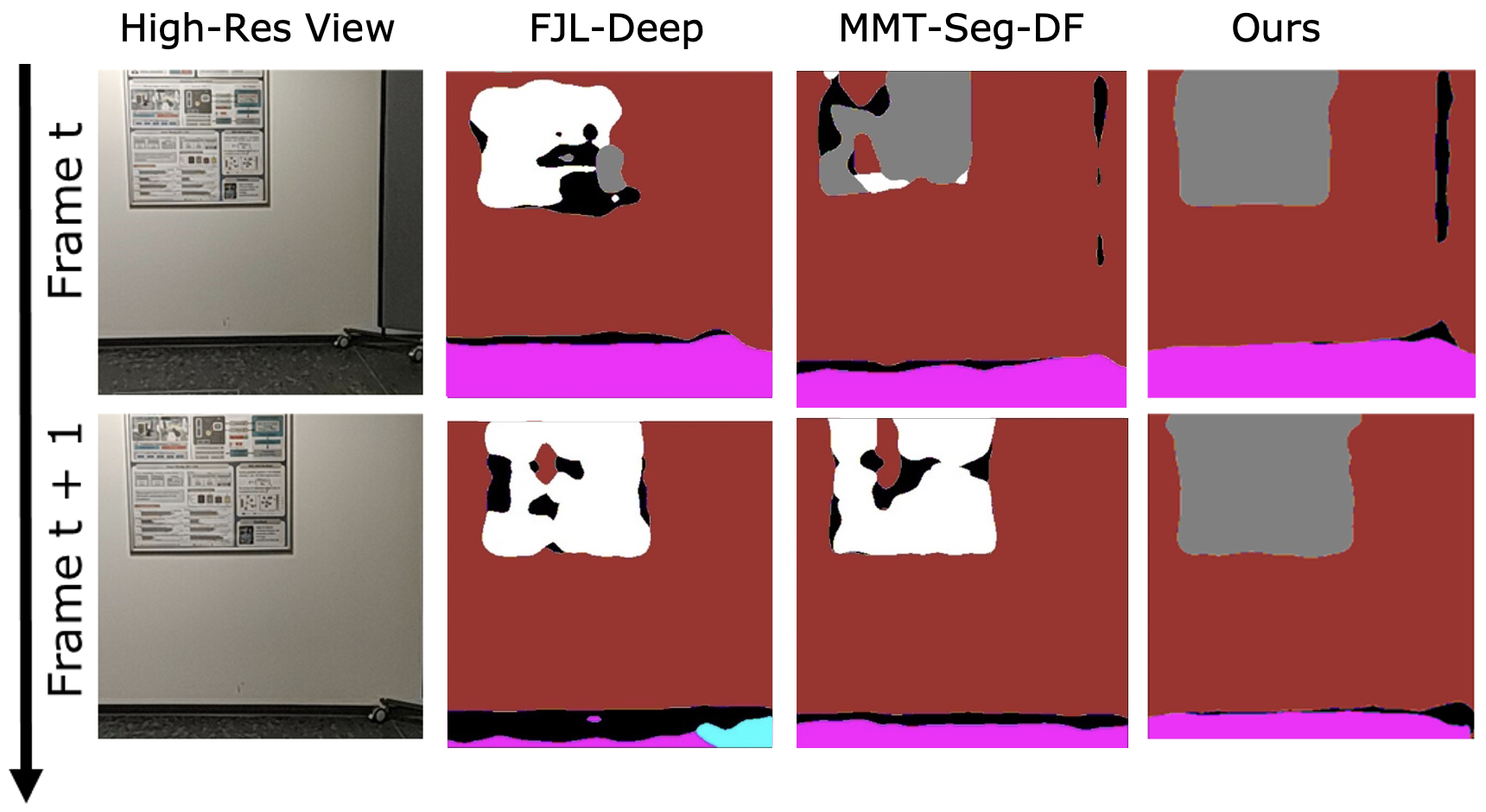}%
  }
  \caption{Semantic segmentation results during navigation across two consecutive frames. The first row shows the high-resolution image for visualization at time step t and its corresponding segmentation outputs from FJL-Deep, MMT-Seg-DF and our method. The second row presents the results for the subsequent frame (t+1). Compared to baselines, our method yields more stable and consistent segmentation across frames, enabling successful navigation by continuously segmenting the target object. In contrast, FJL-Deep, MMT-Seg-DF exhibit segmentation failures in the subsequent frame, which leads to navigation failure to the target object.}
  \label{fig:frame}
  \vspace{-0.3cm} 
\end{figure}

\textbf{Experiment Results:} From the results shown in Table~\ref{table:success_rate}, our method achieves the highest navigation success rate compared to other baselines, outperforming FJL-Deep and MMT-Seg-DF by 22.5 and 20 percentage points respectively. Furthermore, we utilize the full-resolution model, PP-SAM-384 (Full), as a theoretical upper bound. Our approach narrows the performance gap to this upper bound, falling short by only 7.5 percentage points.
Under a privacy-constrained scenario using only ULR RGB images as inputs, our improved ULR semantic segmentation results increase the success rate of the semantic object-goal navigation.
To further illustrate how improved mIoU contributes to improved navigation, we present semantic segmentation results across two consecutive frames in Fig.~\ref{fig:frame}.
These results demonstrate that successful navigation toward a target object requires consistently accurate segmentation across frames.
Our method yields more stable and coherent segmentation results, whereas FJL-Deep and MMT-Seg-DF fail to maintain correct segmentation in consecutive frames, ultimately causing the robot to miss the target.

\section{Discussion} 
\textbf{Boundary Segmentation:} While our method achieves strong ULR semantic segmentation performance, fine-grained boundary segmentation performance remains limited, as indicated by the BF results in Sec.~\ref{sec:SR_SS}.
A plausible explanation is color bleeding around object boundaries, which may weaken contour precision even when region-level semantics are correctly recovered. 
This suggests that future work could benefit from boundary-aware supervision, such as boundary-aware loss or edge-consistency regularization, to better preserve semantic edges and thin structures.

\textbf{ULR Sensing Hardware Validation:} The ULR inputs in our experiments are generated through synthetic downsampling rather than captured by native ULR sensors. Although this synthetic downsampling is a standard protocol~\citep{Ryoo2018aaai} that effectively validates our framework, the deployment realism on native ULR sensing hardware can be further investigated in the future.

\textbf{Privacy-Preserving Multimodal Extension:} Our current method focuses on single-modality RGB inputs. To address the aforementioned boundary degradation and improve task-level robustness, a natural extension is to incorporate privacy-preserving depth sensing~\citep{huang2025arxiv, liu2025arXiv}. Since depth images contain minimal appearance details, they inherently mitigate privacy risks while supplying rich structural information. Exploring this multi-modal fusion could provide complementary geometric cues for boundary localization and enable more complex spatial understanding for downstream tasks such as 3D mapping.

\section{Conclusion}
In this paper, we address the challenge of recovering semantic segmentation from privacy-preserving ultra-low-resolution RGB inputs. We propose a novel joint-learning framework by integrating a Segmentation-Aware Discriminator and an Agglomerative Feature Extractor to mitigate the optimization conflicts between image super-resolution and semantic segmentation exacerbated by severe visual degradation. 
Extensive experiments reveal that effective semantic segmentation should be treated as a task-oriented recovery problem rather than a purely photorealistic reconstruction problem under ultra-low-resolution conditions. 
Our proposed method outperforms diverse representative baselines in semantic segmentation performance and our ultra-low-resolution RGB input achieves a favorable trade-off, balancing privacy preservation and semantic segmentation performance.
Finally, we also deployed our proposed privacy-preserving semantic segmentation method in a real-world robotic object-goal navigation task, showing the successful downstream task execution under severe visual degradation.

\backmatter

\section*{Declarations}

\textbf{Funding:}
This work was partially funded by the German Federal Ministry of Research,
Technology and Space (BMFTR) under grant No.~16KIS1949 and by the Robotics
Institute Germany (RIG).

\textbf{Data availability:}
The \mbox{SUN RGB-D} dataset used in this work is publicly available
from its original source. 
The collected privacy dataset and the real-world indoor dataset are available from the corresponding author on reasonable
request and subject to applicable access conditions.

\textbf{Conflict of interest:}
The authors declare that they have no conflict of interest.

\begin{appendices}

\section{}\label{sec:p_types}
\textbf{Privacy Preservation Results across Categories:} While Sec.~\ref{sec:ULR_threshold} analyzes P-non over the entire dataset, here we present the class-wise results for each sensitive category, as shown in Table~\ref{tab:p_types}.

\begin{table*}[h]
\centering
\renewcommand{\arraystretch}{1.2}
\begin{tabular}{|l|c|c||c|c|c|c|}
\toprule
Prompt & Resolution & Whole & Face & Debit Card & Passport & Private Chat \\
\midrule
\multirow{4}{*}{Original}
& $384\times384$ & 0.0615 & 0.0000 & 0.0714 & 0.0000 & 0.2222 \\
\cmidrule(lr){2-7}
& $32\times32$ & 0.4231 & 0.0698 & 0.3571 & 0.5313 & 0.9259 \\
\cmidrule(lr){2-7}
& $16\times16$ & 0.4769 & 0.3721 & 0.3571 & 0.5625 & 0.6667 \\
\cmidrule(lr){2-7}
& $8\times8$ & 0.6615 & 0.2093 & 0.7143 & 0.9375 & 1.0000 \\
\midrule
\multirow{4}{*}{Readability-Aware}
& $384\times384$ & 0.2692 & 0.1163 & 0.4286 & 0.3750 & 0.2222 \\
\cmidrule(lr){2-7}
& $32\times32$ & 0.6615 & 0.5166 & 0.5357 & 0.6875 & 1.0000 \\
\cmidrule(lr){2-7}
& $16\times16$ & 1.0000 & 1.0000 & 1.0000 & 1.0000 & 1.0000 \\
\cmidrule(lr){2-7}
& $8\times8$ & 1.0000 & 1.0000 & 1.0000 & 1.0000 & 1.0000 \\
\bottomrule
\end{tabular}
\caption{Overall and class-wise privacy non-leakage rate comparison across four different resolutions: $384{\times}384$, $32{\times}32$, $16{\times}16$, and $8{\times}8$, evaluated under both the original and readability-aware prompts.}
\label{tab:p_types}
\end{table*}

A representative example is the \textit{debit card} category. As shown in Fig.~\ref{fig:privacy_visual}, although the card content in the 16$\times$16 image is not visually readable, the original prompt still yields a low P-non value of 0.3571, as reported in Table~\ref{tab:p_types}. 
This suggests that the evaluator may rely on coarse shape cues, such as the overall card-like outline, rather than actual readability. 
In contrast, the readability-aware prompt mitigates this issue by explicitly requiring the model to consider whether the private content is readable. 
As a result, the P-non for the debit card category at 16$\times$16 increases to 1.0000, which is better aligned with the visual observation that the sensitive card details are no longer discernible.


\section{}\label{sec:alpha}
\textbf{Hyperparameter Ablation Study:}
To mitigate the objective conflict between super-resolution and semantic segmentation, we conduct an ablation study on the hyperparameter $\alpha$ as shown in Table~\ref{tab:alpha_ablation}.

\begin{table}[h]
{\centering
\renewcommand{\arraystretch}{1.2}
\resizebox{0.55\columnwidth}{!}{%
\begin{tabular}{|c|c|c|}
\toprule
$\alpha$ & mIoU & mAcc \\
\midrule
0.1 & 0.2775 & 0.3845 \\
\midrule
0.3 & \textbf{0.3101} & \textbf{0.4620} \\
\midrule
0.5 & 0.2678 & 0.3580 \\
\midrule
0.7 & 0.2594 & 0.3403 \\
\bottomrule
\end{tabular}%
}
\par}
\caption{Ablation study on the selection of hyperparameter $\alpha$. We evaluate the performance in terms of mIoU and mAcc. The best performance among the selected variants is achieved when $\alpha = 0.3$.}
\label{tab:alpha_ablation}
\end{table}

Table~\ref{tab:alpha_ablation} shows that properly balancing the super-resolution and semantic segmentation objectives is critical for achieving ULR semantic segmentation. 
The best performance is achieved at $\alpha=0.3$, indicating that this setting most effectively mitigates the optimization conflict between the two objectives among the selected variants.

\section{} \label{sec:nav_system}
\textbf{Semantic Object Goal Navigation System:} 
Algorithm~\ref{alg:algorithm} outlines the overall pipeline of our proposed navigation system. For clarity, we further describe the roles of its main modules in the navigation process as follows.

\begin{algorithm}[h]
\caption{Semantic Object-Goal Navigation}\label{alg:algorithm}
\DontPrintSemicolon
\KwReq{target object $T_o$}

$S_{\mathrm{pred}} \leftarrow \mathrm{SemanticRecovery}()$\;

\While{$T_o \notin S_{\mathrm{pred}}$}{
    $w \leftarrow \mathrm{FloorBasedNavigation}()$\;
    \If{$w = \mathrm{None}$}{
        \Return Target Not Found\;
    }
    $\mathrm{MoveTo}(w)$\;
    $S_{\mathrm{pred}} \leftarrow \mathrm{SemanticRecovery}()$\;
}
$\mathrm{ObjectGoalNavigation}(S_{\mathrm{pred}}, T_o)$\;
\Return Target Found\;
\end{algorithm}

\textbf{SemanticRecovery:} At each waypoint, the robot executes a 360-degree rotation while capturing one ULR image every 36 degrees, yielding 10 images in total. These images are subsequently processed by the proposed ULR semantic segmentation model to determine whether the target object can be identified in the observed scene.

\textbf{ObjectGoalNavigation:} Once the target object is detected, the system switches to the object-goal navigation module. The robot first adjusts its orientation through in-place rotations of 10 degrees until the center of the target object is aligned with the vertical centerline of the image. It then moves toward the target in increments of 20 cm. This process continues until the target object occupies at least 40\% of the image area, at which point the robot stops, indicating the completion of the navigation task.

\textbf{FloorBasedNavigation:} This module is activated when the target object is not detected by \textbf{SemanticRecovery}. Given the current segmentation mask, it first constructs a binary floor map in which only the floor region is treated as foreground. A row-wise scan from the bottom to the top of the image is then performed to extract the midpoint of the floor region in each row, and these midpoints collectively define the floor centerline. 
Waypoints are sampled at fixed intervals along this centerline, and the nearest waypoint~($x_w, y_w$) is selected as the next navigation target. Let~($x_r, y_r$) denote the robot position in the image plane. The robot rotates according to the horizontal offset $\Delta x = x_w - x_r$ to align with the waypoint direction, and then moves forward accordingly. 
To further support exploration, the left and right foreground--background boundaries are fitted with two lines to identify potential branches. 
A region is regarded as a branch towards a new area when foreground pixels deviate from the fitted boundary by more than 35 pixels in the horizontal direction. 
If no valid waypoint is available, the system terminates and reports failure; otherwise, the robot moves to the selected waypoint and repeats the search process.

\end{appendices}


\bibliography{sn-bibliography}

\end{document}